# Machine learning tools to improve nonlinear modeling parameters of RC columns

Hamid Khodadadi Koodiani[a], Elahe Jafari[a], Arsalan Majlesi[a], Mohammad Shahin[b], Adolfo Matamoros[a], Adel Alaeddini[b]

[a] Civil Engineering Department, The University of Texas at San Antonio, San Antonio, USA

[b] Mechanical Engineering Department, The University of Texas at San Antonio, San Antonio, USA

## Abstract

Modeling parameters are essential to the fidelity of nonlinear models of concrete structures subjected to earthquake ground motions, especially when simulating seismic events strong enough to cause collapse. This paper addresses two of the most significant barriers to improving nonlinear modeling provisions in seismic evaluation standards using experimental data sets: identifying the most likely mode of failure of structural components, and implementing data fitting techniques capable of recognizing interdependencies between input parameters and nonlinear relationships between input parameters and model outputs. Machine learning tools in the Scikit-learn and Pytorch libraries were used to calibrate equations and black-box numerical models for nonlinear modeling parameters (MP) a and b of reinforced concrete columns defined in the ASCE 41 and ACI 369.1 standards, and to estimate their most likely mode of failure. It was found that machine learning regression models and machine learning black-boxes were more accurate than current provisions in the ACI 369.1/ASCE 41 Standards. Among the regression models, Regularized Linear Regression was the most accurate for estimating MP a, and Polynomial Regression was the most accurate for estimating MP b. The two black-box models evaluated, namely the Gaussian Process Regression and the Neural Network (NN), provided the most accurate estimates of MPs a and b. The NN model was the most accurate machine learning tool of all evaluated. A multi-class classification tool from the Scikit-learn machine learning library correctly identified column mode of failure with 79% accuracy for rectangular columns and with 81% accuracy for circular columns, a substantial improvement over the classification rules in ASCE 41-13.

*Introduction*

ASCE 41 [1] is a standard for seismic evaluation of building structures widely used in the US. Chapter 10 of ASCE 41 provides guidelines for creating nonlinear models of reinforced concrete structures that replicate the ACI 369.1 [2] Standard with minor changes. Nonlinear analysis procedures in the ACI 369.1/ASCE 41 Standards rely on lateral force versus lateral deformation envelopes, shown in Figure 1, to simulate nonlinear element behavior. Element modeling parameters (*MP*s) and acceptance criteria (*AC*) provided in ACI 369.1/ASCE 41 are crucial components of nonlinear evaluation procedures as they provide an objective basis to construct numerical models and evaluate structural performance. Two critical parameters that define the shape of load-deformation envelopes are nonlinear modeling parameters '*a*' and '*b*', defined as the plastic deformation at incipient lateral-strength degradation (loss of lateral load capacity) and at incipient axial degradation (loss of the ability to carry axial load in columns), respectively. These two parameters are shown in figure 1. Parameter *c* in Figure 1 corresponds to the residual lateral strength and is outside the scope of this study.

Defining modeling parameters for a broad range of elements is challenging because physical behavior is driven by different sets of input variables relevant to the controlling mode for each particular element. Evaluation standards like ACI 369.1/ASCE 41 approach this problem by specifying rules for failure mode classification for each element type (i.e. beams, columns, walls), or by adopting equations that provide the most accurate modeling parameter estimates for all elements of a given type, regardless of the failure mode. The former approach is more precise, but also more complex. Its main disadvantages are that it requires calibrating equations based on statistical analyses of smaller bins of data, that it yields inconsistent estimates of *MP*s across boundaries between failure modes, and that *MP* estimate accuracy near failure mode boundaries is dependent on the accuracy of the failure mode decision tree. The latter approach is simpler, allows calibrating equations based on larger data sets, and provides smooth transitions for estimates of *MP*s across failure mode boundaries, but is less effective in capturing behaviors that are specific to small element subsets, particularly when using linear regression models. Both approaches are hampered by the difficulty inherent in calibrating complex nonlinear relationships between input variables and *MP*s.

As machine learning tools become more pervasive, code committees and building officials will face the decision of whether to allow their use in codes and standards and if so, how to safely implement their use. Evaluating the performance of linear regression models, regression models enhanced with machine learning tools, and machine learning black boxes to estimate *MP*s is of fundamental importance to address this question. Performance metrics that must be considered include accuracy, explainability, and loss of accuracy at the edges of data sets. In the context of this study, machine learning black boxes are defined as algorithms that mimic the human brain and can be viewed in terms of inputs and outputs, without explicit representation to the user of their internal workings.

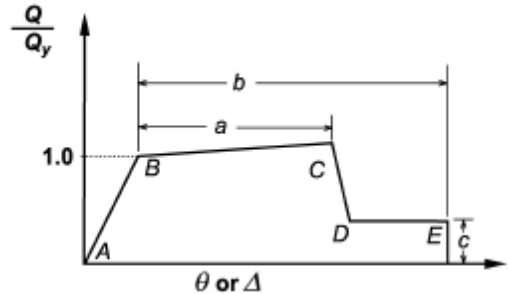

*Figure 1. ASCE 41-06 backbone response for nonlinear modeling of RC components. [3]*

In studies conducted before the early 2000s, it was common practice to refer to the deformation corresponding to a 20% loss of lateral load capacity as the limiting drift or deformation capacity of columns (i.e. [4]). Within the framework of ACI 369.1/ASCE 41, it is common practice to associate this deformation limit with point *C* in the load-deformation envelope curve (Figure 1), and use experimental measurements to calibrate the plastic rotation corresponding to nonlinear parameter *a* [5]. There is a significant body of research on the deformation limit corresponding to point *C* in columns [4, 6-9].

Studies on the deformation limit corresponding to loss of axial load capacity in columns are far fewer [5, 10-14]. Within the framework of ACI 369.1/ASCE 41 modeling parameters, this deformation limit corresponds to point *E* in the load-deformation envelope curve (Figure 1) and experimental measurements are used to calibrate the plastic rotation corresponding to nonlinear parameter *b*.

*MP* and *AC* for reinforced concrete columns in ASCE 41 have changed significantly since the first edition of the standard published in 2006 and are closely related to the expected mode of failure. Numerical values for RC column *MP*s listed in ASCE 41-06 [3], adopted from FEMA 356 [15], made no distinction between rectangular and circular columns and were based primarily on engineering judgment. Column *MP*s in Table 6-8 of ASCE 41-06 were classified into 4 different conditions depending on the expected mode of failure: columns controlled by flexure, shear, inadequate development or splicing length, and columns with axial loads exceeding 0.7 $P_o$. Explicit rules were not provided in column failure classification, leaving engineers to use strength provisions for shear, flexure, and development length in Chapter 6 and engineering judgment to determine the controlling mode of failure. Because the decision tree only considered primary modes of failure, it was not possible to make a distinction between columns controlled by flexure and columns that experienced shear failure after flexural yielding.

Later editions of the ASCE 41 standard placed greater emphasis on providing nonlinear *MP*s based on experimental data, which led to changes both in the magnitude of *MP*s as well as the parameters that define them. *MP*s calibrated using column experimental data were first introduced in Supplement No.1 to ASCE 41 [3], and formally added to Table 10-8 of ASCE 41-13 [16]. *MP* values in Supplement No. 1 and ASCE 41-13 were calibrated to provide conservative estimates of element modeling parameters and made no distinction between rectangular and circular columns. The new *MP*s and *AC* in Table 10-8 of ASCE 41-13 were classified according to conditions *i* to *iv*, corresponding to columns controlled by flexure (*i*), flexure-shear (*ii*), shear (*iii*), and inadequate development or splice length (*iv*). A very simple decision tree to determine column condition (failure mode) for the first three categories was introduced, introducing for the first time a distinction based on the secondary mode of failure between columns controlled by flexure and columns that failed in shear after flexural yielding. Column condition in the decision tree depended on two parameters, the ratio $V_p/V_o$ and the type of hoop anchorage detail.

Several changes to column *MP* provisions were introduced in ASCE 41-17 [1]. The failure mode decision tree was eliminated, consolidating conditions *i*, *ii*, and *iii* in ASCE 41-13 [16] into a single category in ASCE 41-17 with the goal of eliminating inconsistencies across failure mode boundaries. Column *MP*s were defined using equations instead of listing numerical values. Furthermore, equations for column *MP*s were re-calibrated to produce mean estimates instead of conservative estimates of nonlinear response, and different sets of *MP*s were specified in Tables 10-8 and 10-9 for rectangular columns and circular columns with spiral reinforcement or seismic hoops, respectively.

Equations for *MP a* of columns controlled by flexure, flexure-shear, and shear in ACI 369.1-17/ASCE 41-17 are based on a statistical analysis of experimental data by Ghannoum and Matamoros [5]. In their study, they used a linear normal regression model and determined that the most significant variables affecting *MP*s *a* and *b* were axial load ratio $\frac{P}{A_g f'_c}$, transverse reinforcement ratio $\rho_t = \frac{A_v}{b_w s}$, and shear load ratio $\frac{V_y}{V_o}$. Regarding the shear load ratio, it is important to note that Ghannoum and Matamoros used Equation 10-3 of ASCE 41-13 to calculate the nominal shear strength, $V_0$, and not shear strength equations in ACI 318, so the same approach was adopted in this study. The following equations for rectangular and circular column *MP*s were proposed by Ghannoum and Matamoros based on their regression analysis [5]:

$$a_R = 0.042 - 0.043 \frac{P}{f'_c} A_g + 0.063 \rho_t - 0.023 \frac{V_y}{V_o} \geq 0.0 \ (rad) \quad (1)$$

$$b_{2R} = 0.051 - 0.051 \frac{P}{f'_c} A_g + 1.3 \rho_t - 0.023 \frac{V_y}{V_o} \geq a_R \ (rad) \quad (2)$$

$$a_C = 0.06 - 0.058 \frac{P}{f'_c} A_g + 1.3 \rho_t - 0.037 \frac{V_y}{V_o} \geq 0.0 \ (rad) \quad (3)$$

$$b_{2C} = 0.064 - 0.07 \frac{P}{f'_c} A_g + 2.85 \rho_t - 0.03 \frac{V_y}{V_o} \geq a_C \ (rad) \quad (4)$$

Calibrating equations for modeling parameter *b* is difficult because there are very few experiments where rectangular columns were submitted to lateral load reversals until axial failure (36 in the data set used by Ghannoum and Matamoros [5]) and almost none for circular columns (9 in the data set used by Ghannoum and Matamoros [5]). This set of 45 columns where lateral deformation at axial failure was recorded, was designated $b_1$ by Ghannoum and Matamoros. To overcome the problem of having a small data set, Ghannoum and Matamoros generated additional data corresponding to lower bound estimates of MP *b* by considering the maximum inelastic rotation that columns were subjected to if the test loading protocol precluded them from reaching axial failure. The set of lower-bound generated data was designated as $b_2$. Ghannoum and Matamoros used the combined sets $b_1$ and $b_2$ to calibrate Eqs. (2) and (4) through linear regression. The calibration produced larger estimates of *MP* $b_2$, i.e. the regression model was calibrated to produce unconservative estimates of $b_2$, which they considered appropriate given the conservative nature of the generated data. Mean experimental/model $b_1$ ratio calculated using Eqs. (2) and (4) were 1.07 for rectangular columns and 1.64 for circular columns.

Out of concern for the limitations of the generated set $b_2$, Ghannoum and Matamoros [5] also proposed an equation to calculate parameter *b* based on the Elwood and Moehle [11] behavioral model and validated it with the data set $b_1$. The equation for parameter *b* based on the behavioral model was adopted in ASCE 41-17. Because Equation (3) produced estimates of *MP a* larger than those calculated with Equation (1) by a factor of approximately 1.3, Ghannoum and Matamoros [5] recommended increasing estimates of *MP b* obtained with behavioral model equations for rectangular columns by 30% to account for the improved performance of spirally reinforced circular columns.

The main limitation of linear normal regression models such as the one used by Ghannoum and Matamoros [5] is that when applied to a broad data set like the column database used in this study the most important variables vary for data subsets depending on the mode of failure. This is illustrated in Table 1, which presents the results from a P-Value analysis for different rectangular column bins of the complete set considered in this study. Bins in Table 1 were defined in terms of the shear span-to-depth ratio (short, intermediate, and slender columns), axial load ratio (approximately beams, tension-controlled columns, and compression-controlled columns), and failure mode. Table 1 clearly shows that the subset of most relevant input variables was different for each bin. Furthermore, a similar analysis for circular columns yields entirely different sets of relevant variables for each bin. The accuracy achieved by using a single set of input variables for the entire set was highest for tension-controlled columns with $R^2$ of 43% for *MP a* and 21% for *MP b*. Accuracy improved significantly when different sets of equations were calibrated for each bin, particularly for bins with unique behavior characteristics like short columns ($R^2$ of 60% for *MP a*) and shear critical columns ($R^2$ of 100% for *MP a* and 60% for *MP b*). In general, most bins achieved $R^2$ values ranging between 20 and 40%.

Table 1. P-Value analysis for different subsets of rectangular column.

| Range | $\frac{a}{d} < 3$ | | $3 \leq \frac{a}{d} \leq 5$ | | $\frac{a}{d} > 5$ | |
|---|---|---|---|---|---|---|
| MP | a | b | a | b | a | b |
| $R^2$ (%) * | 43, 60 | 13, 26 | 32, 52 | 39, 48 | -9, 43 | 6, 46 |
| MSE x $10^5$ | 17, 12 | 45, 38 | 22, 16 | 28, 23 | 30, 16 | 40, 23 |
| Most significant variables | $\rho_t$ | $\rho_t$ | $\frac{a}{d}$ | $\frac{P}{f'_c}A_g$ | $\frac{s}{d}$ | $\frac{s}{d}$ |
| | $\frac{P}{f'_c}A_g$ | $\frac{P}{f'_c}A_g$ | $\frac{P}{f'_c}A_g$ | $\frac{V_y}{V_o}$ | $\rho_l$ | $\rho_l$ |
| | $\frac{a}{d}$ | $\frac{a}{d}$ | $\frac{V_y}{V_o}$ | $\rho_l$ | $\rho_t$ | $\frac{P}{f'_c}A_g$ |
| Range | $\frac{P}{f'_c}A_g < 0.1$ | | $0.1 \leq \frac{P}{f'_c}A_g \leq 0.3$ | | $\frac{P}{f'_c}A_g > 0.3$ | |

| MP | a | b | a | b | a | b |
|---|---|---|---|---|---|---|
| $R^2$ (%) * | 37, 48 | 19, 32 | 43, 44 | 21, 35 | 30, 30 | 41, 45 |
| MSE x $10^5$ | 24, 20 | 33, 28 | 21, 20 | 46, 37 | 22, 22 | 26, 24 |
| Most significant variables | $\frac{V_y}{V_o}$ | $\frac{V_y}{V_o}$ | $\frac{V_y}{V_o}$ | $\frac{s}{d}$ | $\rho_l$ | $\rho_l$ |
| | $\rho_l$ | $\frac{a}{d}$ | $\frac{P}{f'_c}A_g$ | $\rho_l$ | $\frac{V_y}{V_o}$ | $\frac{V_y}{V_o}$ |
| | $\frac{a}{d}$ | $\rho_l$ | $\rho_l$ | $\frac{V_y}{V_o}$ | $\frac{P}{f'_c}A_g$ | $\frac{V_y}{V_o}$ |
| **Failure** | **Flexure Critical** | | **Flexure-Shear Critical** | | **Shear Critical** | |
| MP | a | b | a | b | a | b |
| $R^2$ (%) * | 17, 28 | 23, 24 | 23, 36 | 23, 35 | -, 100 | -1, 13 |
| MSE x $10^5$ | 27, 24 | 38, 37 | 13, 11 | 17, 14 | 14, 0 | 76, 65 |
| Most significant variables | $\frac{P}{f'_c}A_g$ | $\frac{P}{f'_c}A_g$ | $\frac{P}{f'_c}A_g$ | $\frac{P}{f'_c}A_g$ | - | $\frac{s}{d}$ |
| | $\frac{s}{d}$ | $\rho_l$ | $\rho_t$ | $\rho_l$ | - | $\rho_t$ |
| | $\rho_l$ | $\frac{s}{d}$ | $\frac{a}{d}$ | $\rho_t$ | - | $\rho_l$ |

*\* All regression equations were calibrated with the three most significant variables of the corresponding bin. The first number corresponds to equations calibrated and applied to the complete column data set and the second corresponds to equations calibrated and applied only to data in the bin.*

*Literature Review*

In structural engineering, machine learning (ML) tools have been extensively studied in recent years. A wide range of structural engineering applications, such as structural analysis, design, health monitoring, element mechanical behavior and capacity, optimization, and others, make use of ML tools. Thai (2022) reviewed a variety of ML applications in structural engineering [17]. The compressive strength of fabric-reinforced cementitious matrix (FRCM)-reinforced concrete columns was predicted using a neural network (NN) model by Irandegani et al. in 2022 [18]. In order to predict the displacement ductility ratio of RC joints in 2022, Dabiri et al. applied NN, Random Forest, and regression-based models [19]. The compressive strength of FRP-confined reinforced concrete columns was calculated using the second-order polynomial Response Surface

Methodology (RSM) by Moodi et al. (2018) [20]. Huan Luo et al. (2021) proposed "a nature-inspired, metaheuristic, least squares-support vector machine for regression (LS-SVMR) model for enhancing the generalization of lateral strength prediction of RC columns" [21]. For predicting the backbone curves for flexure- and shear-critical columns, Huan Luo and Stephanie German Paal (2018) developed a novel machine-learning-based backbone curve model (ML-BCV). In comparison to conventional modeling methodologies, the proposed model could reduce error [22]. In 2022 Huang et al. used a support vector machine (SVM) to classify the different failure modes (i.e., flexure failure, flexure-shear failure, and shear failure). They trained an artificial neural network (ANN) to define the relationship between lateral force and displacement of reinforced concrete columns with different proporties using a database of 498 experimental results [23]. Chaabene et al. (2020) investigated different ML models to predict the compressive, shear and tensile strength, and elastic modulus of concrete [24]. These models included artificial neural networks, support vector machines, decision trees, and evolutionary algorithms. Using 254 cyclic loading tests of RC columns to predict the failure mode and bearing capacity of columns, Feng et al. developed the adaptive boosting (AdaBoost) method [25].

*Research Significance*

This paper addresses a gap in the literature given the paucity of references where machine learning tools have been used to study nonlinear modeling of reinforced concrete columns.

The main objective of this study is to develop equations for columns *MP*s *a* and *b*, as well as equations to classify columns in terms of mode of failure, using machine learning algorithms and tools to improve accuracy. These tools have the ability to capture highly nonlinear relationships between input variables and modeling parameters, which provides the capability of emphasizing the subset of input variables most relevant to the mode of failure for multiple subsets.

Past studies have evaluated the use of machine learning algorithms for fitting highly nonlinear data sets. This study does so in the context of code provisions closely related to an experimental data set that covers multiple modes of failure.

*Data analytics*

490 pseudo-static tests of concrete columns (319 rectangular columns and 171 circular columns) are used in this study. A description of the database can be found in [26]. The database included columns that failed in flexure, flexure followed by shear, and shear. The number of experiments of columns deformed until axial collapse is very small (45 out of 490 in the data set), which makes it impractical to use machine learning tools, so parameter *b* was calibrated considering dataset b2 generated by [5]. The experimental data set used in this study included six nondimensional input variables and two output MPs, *a* and *b*, for each column test. The input variables are column span-to-depth ratio ($\frac{a}{d}$), axial load ratio ($\frac{P}{A_g f'_c}$), longitudinal reinforcement ratio ($\rho_l$), transverse reinforcement ratio ($\rho_t = \frac{A_v}{b_w s}$), transverse reinforcement spacing-to-effective depth ratio ($\frac{S}{d}$), and the ratio of shear demand at yielding of the longitudinal reinforcement-to-shear capacity or shear capacity ratio ($\frac{V_y}{V_0}$). Histograms of input variables used in this study are shown in Figure 2a and 2b for rectangular and circular columns, respectively.

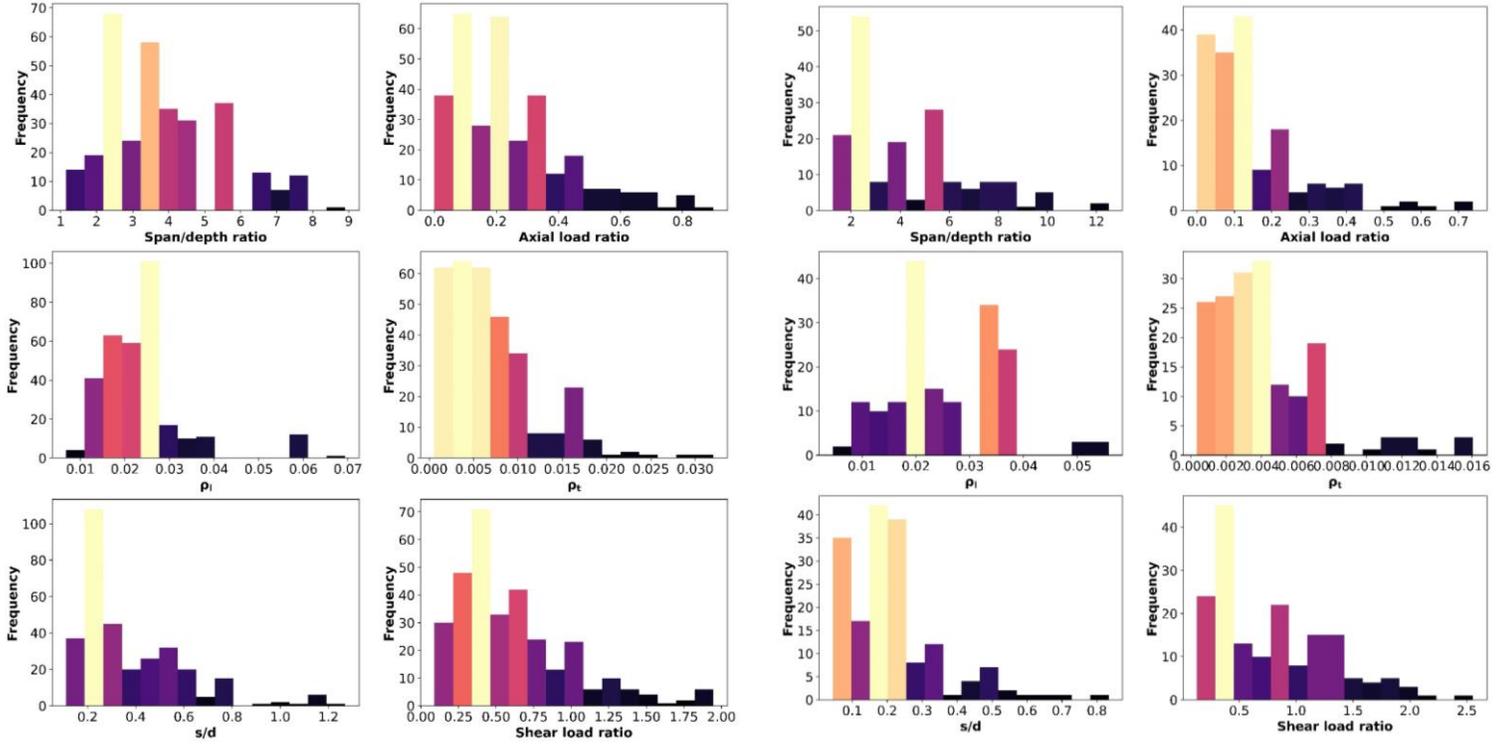

(a) (b)

*Figure 2. Histogram of input variables in column database for a) rectangular and b) circular columns.*

## Data visualization

The six features in the dataset were reduced to two features using the TSNE algorithm [27] in order to facilitate visualization in 2D space. The nonlinear separability of data clusters was then visualized using the Gaussian Mixture Model (GMM) [28]. Using these methods, seven distinct data clusters were identified; the centers of each cluster are depicted in Fig. 3 with a red dot. Individual data points are represented by the black dots in Figure 3, and the clouds around each cluster's center display the distribution of the data. Similar background colors on different parts of Fig. 3 demonstrate linear relationships between the data, while background color variations show nonlinear behavior. If a single equation is used to estimate the MPs of columns in all seven clusters, as shown in Figure 3, there is a significant degree of nonlinearity in the column dataset, making the employment of nonlinear models necessary to accurately reflect the relationships between inputs and outputs.

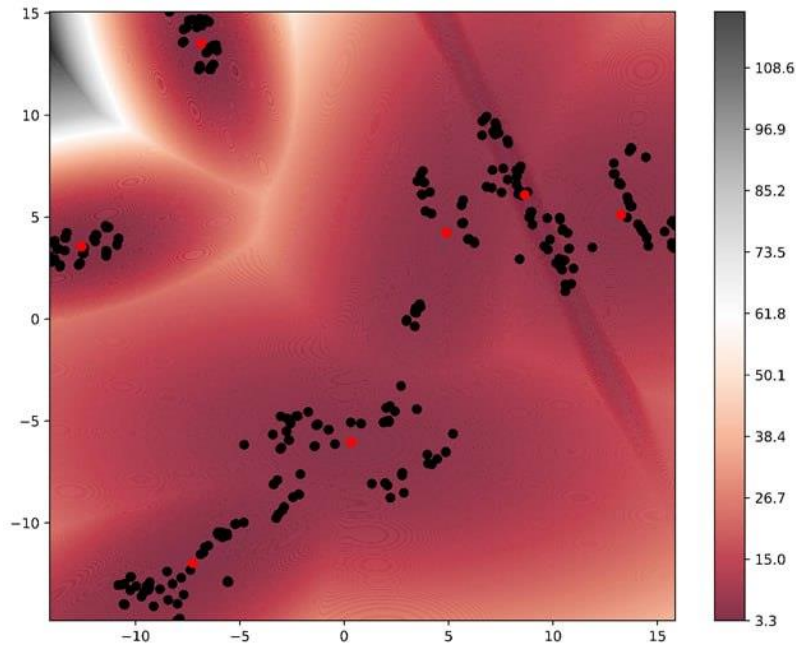

*Figure 3: Negative log-likelihood predicted by GMM for rectangular columns*

**Methodology**

The machine learning methods used in this study are considered supervised learning methods because estimates of modeling parameters are based on experimental results.

Input variables considered in this study were chosen based on behavioral models and experimental observations of column failure mechanisms found in the literature as well as the set of parameters reported in experimental studies, grouped to form a set of non-dimensional parameters. Because all input variables are dimensionless, scaling was not a concern in the configuration of the machine learning models.

To avoid skewing the results of nonlinear simulations, it is important that the proposed MP *a* and *b* follow the same approach adopted in ASCE 41-17 and target the median of test values. The goodness of fit of the proposed models was compared with the goodness of fit of the ASCE 41-06 and ASCE 41-17 Standard equations in terms of two metrics: the standard deviation of the errors, which is a measure of the variance of the results, and R-squared ($R^2$ %), which is a measure of accuracy of the results. Higher values of R-squared are indicative of a better fit between model and experimental observations.

All methods evaluated in this study are implemented in the scikit-learn machine learning library for Python [29]. This library was chosen because it is widely available and easy to implement. The first set of tools evaluated focused on improving linear regression using the same number of variables used by Ghannoum and Matamoros [5] selected before calibration based on their statistical significance. A Multiple Linear Regression (MLR) model was created as a reference. Regularized Linear Regression (RLR) was implemented to evaluate improvement in accuracy obtained by fitting a linear regression model involving all the variables as a first step, followed by calibration of a second model containing only the significant features. In both models, the estimated coefficients are shrunken with an optimum regularization parameter $\lambda$. The benefits of improved sampling were evaluated by implementing the K-fold cross-validation method to fit the best model to the most significant predictors.

A Polynomial Regression Model (PRM) containing the squared form of the most significant variables or features was created to evaluate the improvement in accuracy associated with introducing nonlinear relationships between input variables and *MP*s, without changing the number of input variables.

The last two methods evaluated consisted of numerical tools that produce estimates of modeling parameters given a set of inputs. The first numerical fitting tool was Gaussian Process Regression (GPR), a non-parametric supervised learning method, in which a squared exponential kernel was employed to create new features based on the similarity of the inputs and outputs. The second numerical method evaluated consisted of a Neural Network (NN) trained to recognize the underlying relationships in the dataset and estimate modeling parameters using all input variables in the set. The One-vs-All approach in multiclass classification was used to predict the failure mode of new observations, and Gradient Descent was used to minimize the cost function. Calibration of the NN to calculate modeling parameters of circular columns proved to be more difficult due to the lower number of data available. This problem manifested itself through a larger error in unconservative estimates of *MP*s $a$ and $b$. The performance of NN models can be improved through several different measures, such as, finding more data, creating artificial data, creating new features, optimizing the hyperparameters of the model, and using a deep kernel machine. In this study, because of the simplicity and effectiveness of the process, newly created features and improved optimization of the hyperparameters of the NN were used to improve the accuracy of *MP* estimates for circular columns.

*Multiple Linear Regression (MLR) Model*

The MLR model was calibrated based on the three most significant variables (best features) of the data set. Feature selection was based on their P-Value hypothesis testing to eliminate features that had no relevance to the response parameter. Smaller P-Values are indicative that the corresponding feature is more relevant.

Table 2 shows the P-Values for modeling parameters a and b as outputs calculated using Scikit-learn multiple linear regression. The most significant variables in each case correspond to the shaded cells. Calculated P-Values indicate that axial load ratio, longitudinal reinforcement ratio, and shear load ratio were the most significant variables for estimating MPs *a* and *b* for rectangular columns, and MP *b* for circular columns. Axial load ratio, transverse reinforcement ratio, and shear span-to-depth ratio were the most significant variables for MP *a* of circular columns.

Table 2. P-Values analysis for *MP*s *a* and *b* in circular and rectangular columns.

| Features | Rectangular column P-Values | | Circular column P-Values | |
|---|---|---|---|---|
| | *a* | *b* | *a* | *b* |
| $\frac{a}{d}$ | 0.7285 | 0.2042 | 0.0002 | 0.1975 |
| $\frac{P}{f'_c}A_g$ | 0.0000 | 0.0000 | 0.0000 | 0.0000 |
| $\rho_L$ | 0.0001 | 0.0001 | 0.1551 | 0.0104 |
| $\rho_t$ | 0.0590 | 0.8931 | 0.0001 | 0.2601 |
| s/d | 0.0047 | 0.8345 | 0.0055 | 0.0109 |
| $\frac{V_y}{V_o}$ | 0.0001 | 0.0000 | 0.0035 | 0.0000 |

The MLR model was calibrated using the three features with the most influence on the outputs. Equations for the MLR models of rectangular and circular columns are presented in Eqs. 5 to 8.

$$a_R = 0.046 - 0.043 \frac{P}{A_g f'_c} + 0.363 \, \rho_l - 0.031 \frac{V_y}{V_o} \qquad (5)$$

$$b_R = 0.054 - 0.047 \frac{P}{A_g f'_c} + 0.565 \, \rho_l - 0.03 \frac{V_y}{V_o} \qquad (6)$$

$$a_C = -0.002 - 0.059 \frac{P}{A_g f'_c} + 3.282 \, \rho_t + 0.007 \frac{a}{d} \qquad (7)$$

$$b_C = 0.069 - 0.072 \frac{P}{A_g f'_c} + 0.742 \, \rho_l - 0.044 \frac{V_y}{V_o} \qquad (8)$$

The main difference between Eq. 1 to 4 by Ghannoum and Matamoros [5] is the input parameters. The magnitude of the coefficients for variables that appear in both models such as $\frac{P}{A_g f'_c}$ and $\frac{V_y}{V_o}$ was similar, with the exception of the coefficient for the variable $\rho_t$ for circular columns, which was 252% higher in Eq. 7 than in Eq. 3. Although the variable $\rho_t$ does not appear in Eqs. 5, 6, and 8, shear strength $V_o$ is proportional to $\rho_t$ so there is some degree of cross-correlation between the two.

Figure 5 shows CDFs of error in estimates of *MP*s *a* and *b* calculated with the MLR model, the Ghannoum and Matamoros model [5] (adopted in ASCE 41-17 for MP *a [1]*), and Table 6-8 of ASCE 41-06, which made no distinction between circular and rectangular columns. Equations 5 to 8 were more accurate (had a smaller error) in conservative estimates of *MP*s (error > 0), and larger errors for unconservative estimates of *MP a* (error < 0).

Table 4 shows that for rectangular columns $R^2$ of the MLR model increased by 15 % for *MP a* and by 21% for *MP b*, with respect to Eq. 1 and 2. In the case of circular columns, $R^2$ increased by 14% for MP *a*, and 2% for MP *b* with respect to Eq. 3 and 4.

*Polynomial Regression Model (PRM)*

New features in the form of the square of each significant variable were added to the MLR model to improve precision by avoiding underfitting. The general form of the polynomial equations for *MP*s *a* and *b* is presented in Eq. 9, which had the same three most significant features for each type of column as in Eqs. 5 to 8. Regression coefficients of the model for each column category are presented in Table 3. The standard deviation of the error $R^2$ values of the PRM model for each column category are shown in Table 4.

$$a_{R,C}(b_{R,C}) = \beta_0 + \beta_1 \left(\frac{P}{f'_c} A_g\right) + \beta_2 (\rho_l \text{ or } \rho_t) + \beta_3 \left(\frac{V_y}{V_o} \text{ or } \frac{a}{d}\right) + \beta_4 \left(\frac{P}{f'_c} A_g\right)^2 \\ + \beta_5 (\rho_l \text{ or } \rho_t)^2 + \beta_6 \left(\frac{V_y}{V_o} \text{ or } \frac{a}{d}\right)^2 \quad (9)$$

Table 3. Regression coefficients for PRM model.

| MP | $\beta_0$ | $\beta_1$ | $\beta_2$ | $\beta_3$ | $\beta_4$ | $\beta_5$ | $\beta_6$ |
|---|---|---|---|---|---|---|---|
| $a_R$ | 0.030 | -0.039 | 1.488 | -0.031 | -0.009 | -16.166 | -0.001 |
| $b_R$ | 0.033 | -0.012 | 2.150 | -0.044 | -0.056 | -23.141 | 0.007 |
| $a_C$ | -0.018 | -0.027 | 6.933 | 0.010 | -0.057 | -280.136 | 0.000 |
| $b_C$ | 0.079 | 0.008 | 0.935 | -0.088 | -0.141 | -8.469 | 0.024 |

The CDF of the error in estimates of *MP*s *a* and *b* for the PRM model, the Ghannoum and Matamoros model (2014) (adopted in ASCE 41-17 for MP *a*), and Table 6-8 of ASCE 41-06 are presented in Figure 5. Table 4 shows that for rectangular columns $R^2$ of the PRM model increased by 21 % for MP *a* and by 38% for MP *b*, with respect to Eq. 1 and 2. In the case of circular columns, $R^2$ increased by 20% for MP *a*, and 11% for MP *b* with respect to Eq. 3 and 4. The performance of the PRM model was better than the MLR model in terms of the $R^2$ metric, but the improvement came at the cost of more complex equations and loss of interpretability. Although the performance was better, Figure 5 shows that the PRM model also performed better for conservative estimates of *MP*s *a* and *b* (Error > 0), and performed worse than the ASCE 41 equations for unconservative estimates of *MP an* (Error < 0).

*Regularized Linear Regression (RLR)*

The RLR method avoids underfitting or overfitting by adding a tuning parameter λ to the cost function. To find the optimal tuning parameter, the data set is divided into training (70% of the data points) and validating (30% of the data points) subsets. The relationship between the cost function and the tuning parameter is calculated for the training and validating subsets, and the optimum value of *λ* is chosen so that the added cost of training and validating subsets is minimized. The relationship between the tuning parameter *λ* and cost functions for the error of $b_R$ are presented in Figure 4, for the training and validation subsets. Optimum tuning parameters for $a_R$, $b_R$, $a_C$, and $b_C$ were 0, 1.52, 0, and 2.42, respectively.

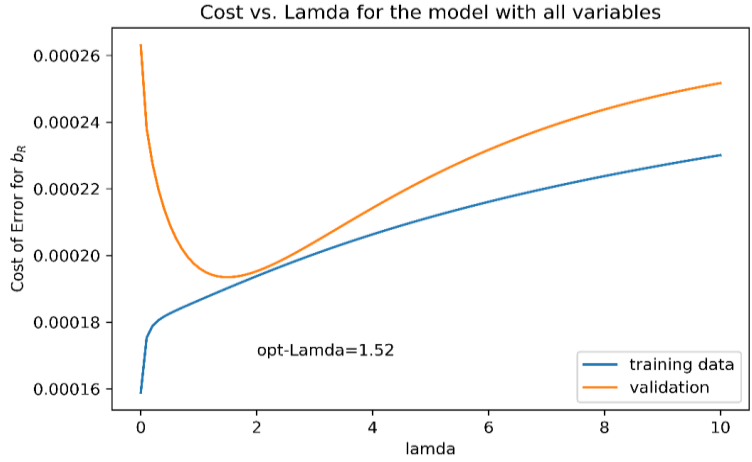

*Figure 4. Relationship between cost function and tuning parameter for training and validation subsets for MP b of rectangular columns.*

The regularized linear regression model results in Eqs. 10 to 13 for *MP*s *a* and *b* of rectangular and circular columns.

$$a_R = 0.052 - 0.0012\frac{a}{d} - 0.046\frac{P}{f'_c}A_g + 0.36\,\rho_l + 0.21\,\rho_t + 0.0074\frac{s}{d} - 0.030\frac{V_y}{V_o} \quad (10)$$

$$b_R = 0.055 + 0.0019\frac{a}{d} - 0.031\frac{P}{f'_c}A_g + 0.01\,\rho_l + 0.0034\,\rho_t - 0.027\frac{s}{d} - 0.012\frac{V_y}{V_o} \quad (11)$$

$$a_C = 0.047 + 0.003\frac{a}{d} - 0.062\frac{P}{f'_c}A_g + 0.440\,\rho_l + 0.622\,\rho_t - 0.031\frac{s}{d} - 0.030\frac{V_y}{V_o} \quad (12)$$

$$b_C = 0.043 + 0.004\frac{a}{d} - 0.022\frac{P}{f'_c}A_g + 0.003\,\rho_l + 0.001\,\rho_t - 0.024\frac{s}{d} - 0.014\frac{V_y}{V_o} \quad (13)$$

Equations 10 to 13 have the advantage that they were derived based on a larger set of input variables that can be representative of different modes of failure. The main drawback is that the coefficients provide linear relationships between *MP*s and each of the input variables, lacking the ability to emphasize and de-emphasize input variables in different ranges of the data set that can only be implemented through the use of nonlinear relationships.

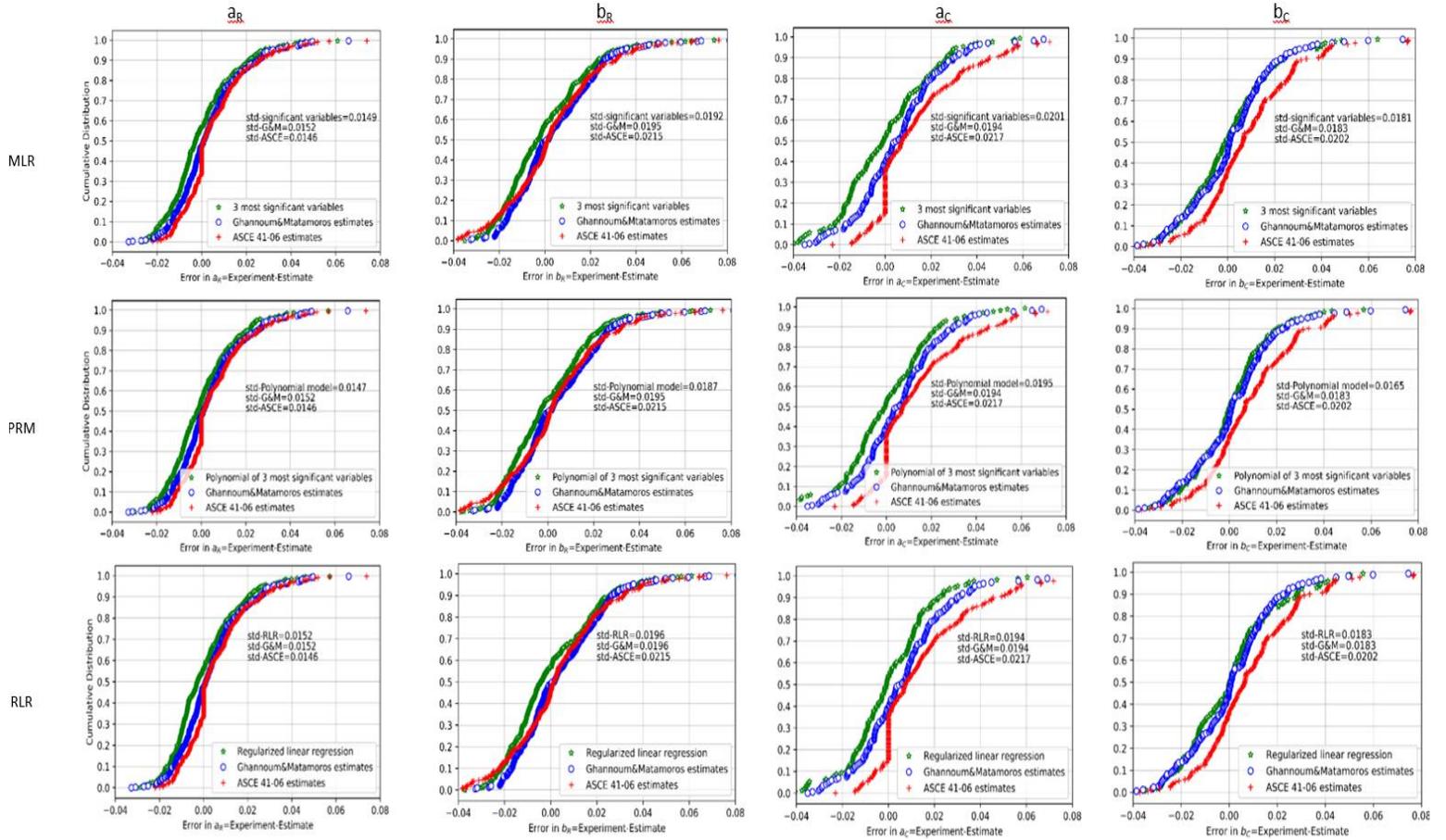

*Figure 5. Cumulative distribution of error for MLR, PRM and RLR models with 3 most significant variables, Ghannoum and Matamoros model, and ASCE 41-06 for $a_R$, $b_R$, $a_c$, and $b_c$.*

Table 4 shows $R^2$ and the standard deviation of the errors for the RLR model. $R^2$ of the RLR model increased by 27% for *MP a* and by 13% for *MP b*, with respect to Eq. 1 and 2. In the case of circular columns, $R^2$ increased by 26% for *MP a*, and decreased by 7% for MP *b* with respect to Eq. 3 and 4. The RLR model performed better than the MLR and PRM models for estimates of *MP a*, but not nearly as well for *MP b* which had much fewer data available for calibration. CDF of the error for the RLR model is plotted in Figure 5. The observed trends were similar to those for the MLR and PRM shown in Figure 5.

*Gaussian Process Regression (GPR)*

Gaussian Process Regression (GPR) is a generic supervised learning method in which estimates are obtained by interpolating observations using the Gaussian kernel function. Kernels use the function being learned by defining the similarity of two data points and assuming that similar data points should have similar target values. The squared exponential kernel, described by Eq. 14, is one of the most commonly used kernels and was used in this study.

$$K(x, x') = \sigma_f^2 \exp\left(\frac{-(x-x')^2}{2\sigma^2}\right) \tag{14}$$

Because *MP*s *a* and *b* exhibit different behaviors in different ranges of input variables, test results do not show a consistent pattern for all columns in the database. For this reason, finding a pattern applicable to the entire set using linear regression models is not possible. This problem is illustrated in Figure 6, which shows the relationship between *MP a* of rectangular columns and axial load ratio, which the P-Value analysis summarized in Table 2 showed to be one of the most significant variables. Data in Figure 6 shows the difficulty of expressing a pattern with a linear or even nonlinear relationship. A significant part of the scatter in Figure 6 originates from the effect of other variables. Figure 6 shows that for each value of axial load ratio, there is a range of different values for *MP a*. This variation is due in part to experimental error, but mostly to the effect of other input variables, likely to be nonlinear in nature. Complex nonlinear interactions among all input variables make the whole system highly nonlinear .

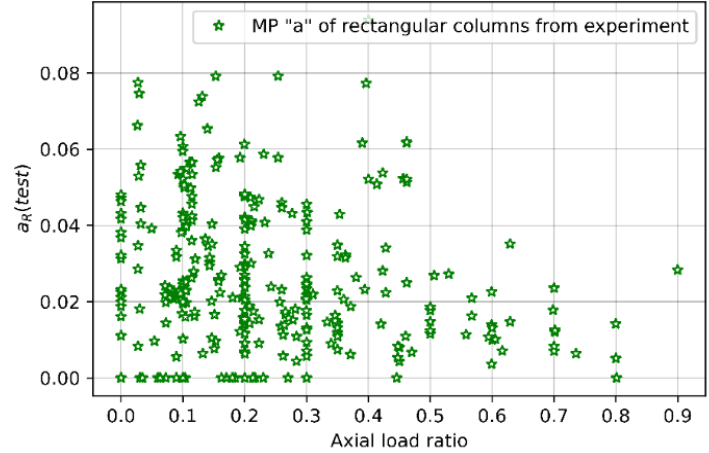

*Figure 6. MP a inferred from experimental data versus axial load ratio for rectangular columns.*

All six input variables were considered in the GRP model. The accuracy and precision of the model were checked by using 90% of the data to train and 10% of the data as the test subset. The cumulative distribution of errors for both *MP a* and *b* in circular and rectangular columns are plotted for the Gaussian process method versus Ghannoum and Matamoros estimates and ASCE estimates in Figure 7. The GP model provides better estimates than other models in terms of the standard deviation and $R^2$.

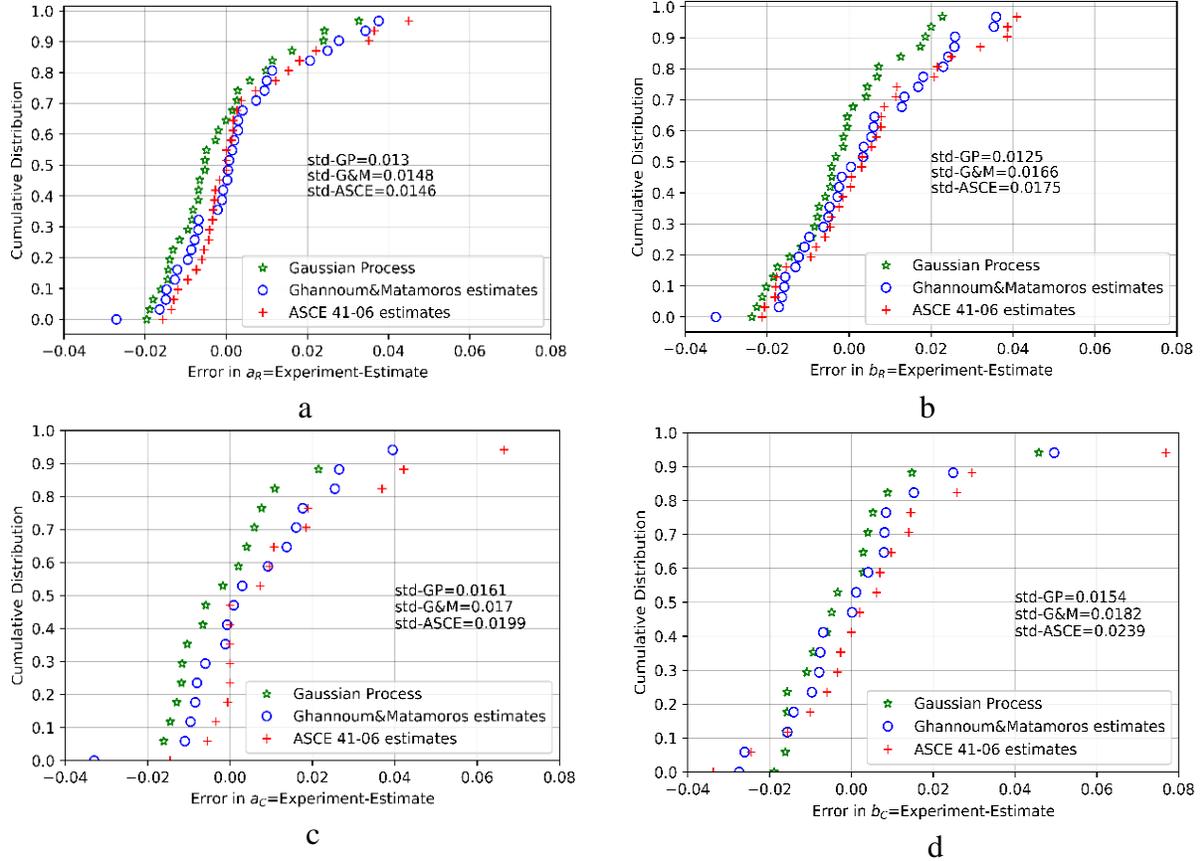

*Figure 7. Cumulative distribution of error for GPR model, Ghannoum and Matamoros model, and ASCE 41-06 for (a) $a_R$, (b) $b_R$, (c) $a_c$, and (d) $b_c$.*

*Neural Network (NN)*

The last method evaluated was the deep Neural Network model (NN) built using the PyTorch library [30] for Python [31]. The accuracy of the NN models is sensitive to the number of iterations (epochs), learning rate, and network configuration, specifically the number of hidden layers and neurons in each layer. Model accuracy increases significantly by increasing the number of neurons, epochs, and hidden layers, but increasing those network configuration parameters also increases computational cost. Model accuracy is also sensitive to the type of activation function used.

The NN used in this study had four hidden layers, 200 neurons, and 10000 iterations, and used the RELU activation function. Test data were divided into a training subset, with 70 percent of the data selected at random, and the validation subset, with the remaining 30 %. This model provided the best results of all models evaluated. For example, model accuracy for *MP a* of rectangular columns was 78 % for the test subset. The CFD of the errors for the NN model, the ASCE 41-06 and ASCE 41-17 provisions are shown in Figure 8. This was the only model evaluated which had smaller errors for conservative and unconservative estimates of MPs *a* and *b*. R-squared for NN models of *MP*s *a* and *b* were 70% and 76% for rectangular columns, and 70% and 66% for the circular columns, respectively. For rectangular columns that represent improvements of 112% and 217% for *MP*s *a* and *b* with respect to the provisions in ASCE 41-17. For circular columns, improvements were 60% and 56% *MP*s *a* and *b*.

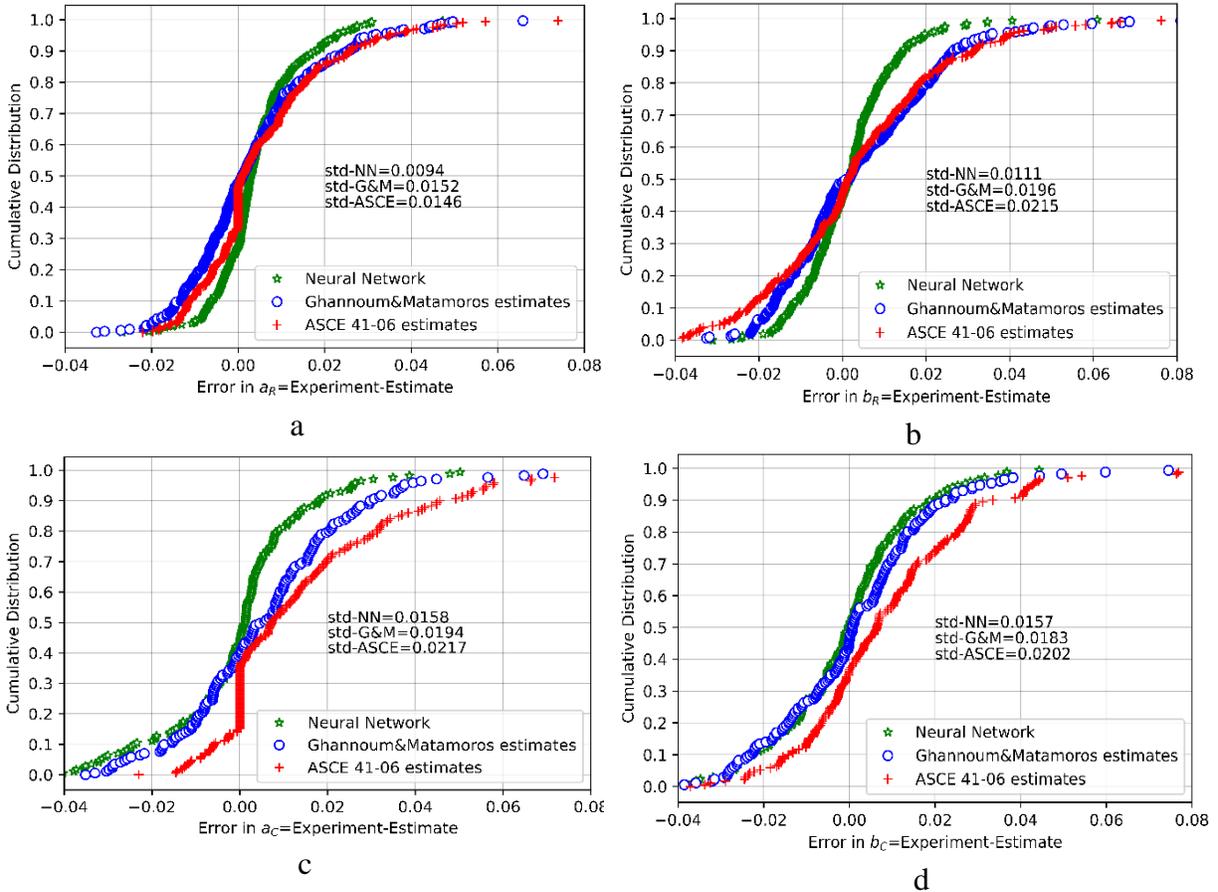

*Figure 8. Cumulative distribution of error for NN model, Ghannoum and Matamoros model, and ASCE 41-06 for (a) $a_R$, (b) $b_R$, (c) $a_c$, and (d) $b_c$.*

Figure 8 shows that unconservative estimates of *MP*s for circular columns obtained with the NN model, shown at the left tail of the CDF curves, had larger errors than the ASCE 41 equations. Several approaches were adopted to reduce the error for this subset of elements. The hyperparameters of the NN model were optimized, and a dynamic learning rate was used with the *torch.optim* module and learning rate scheduler. [32]

In addition, new features are added as input variables to train the NN model based on a parametric variation study of the NN model. The new features, such as the second-order span-to-depth ratio and the second-order shear load ratio, caused the most nonlinearity in the parametric variation study. Knowing these relationships and the effect of each input variable helped to significantly improve the accuracy and precision of the NN model. R-squared increased with the addition of new features to 80% and 84%, for *MP*s *a* and b of circular columns, respectively. The enhanced model for circular columns had much lower errors for unconservative estimates of *MP*s *a* and *b* (left tail of the CDF curves), as shown in Figure 9.

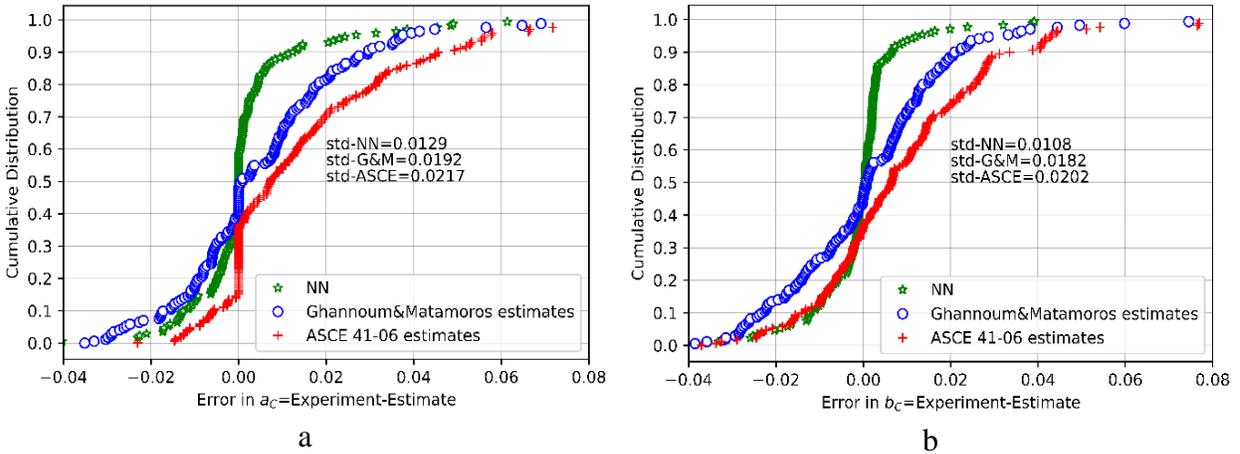

*Figure 9. Cumulative distribution of error for enhanced circular column NN model, Ghannoum and Matamoros model, and ASCE 41-06 for (a) $a_c$, and (b) $b_c$.*

The main limitation of this method is that it is meant to be applied numerically. To facilitate access, a Graphical User Interface (GUI) was added using Python and the model was submitted to GitHub [33] for free distribution. Also, a web service was designed using the google app engine [34] to eliminate the need to use Python to run the model. The cloud implementation of the model provides estimates of *MP*s based on the NN and ASCE 41-17 equations, as well as the expected mode of failure based on the Multinomial Logistic Regression described in a later Section.

*Comparison of machine learning models*

Table 4 lists the Standard Deviation of the errors and $R^2$ for all the models evaluated as well as the Ghannoum and Matamoros regression model and the ASCE 41 Standards.

*Table 4. Statistical parameters for models evaluated in the study.*

| | Rectangular Columns | | | | | | Circular Columns | | | | | |
| --- | --- | --- | --- | --- | --- | --- | --- | --- | --- | --- | --- | --- |
| | *a* | | | *b* | | | *a* | | | *b* | | |
| | $R^2$ (%) | Std. | MSE x $10^5$ | $R^2$ (%) | Std. | MSE x $10^5$ | $R^2$ (%) | Std. | MSE x $10^5$ | $R^2$ (%) | Std. | MSE x $10^5$ |
| MLR | 38 | 0.0149 | 22 | 29 | 0.0192 | 37 | 57 | 0.0186 | 35 | 55 | 0.0181 | 33 |
| PRM | 40 | 0.0146 | 21 | 33 | 0.0187 | 35 | 60 | 0.0181 | 33 | 60 | 0.0170 | 29 |
| RLR | 42 | 0.0145 | 21 | 27 | 0.0196 | 38 | 63 | 0.0175 | 31 | 50 | 0.0190 | 36 |
| GPR | 41 | 0.0130 | 15 | 56 | 0.0125 | 21 | 65 | 0.0161 | 20 | 67 | 0.0154 | 28 |
| NN | 70 | 0.0094 | 10 | 76 | 0.0111 | 12 | 80 | 0.0129 | 16 | 84 | 0.0108 | 11 |
| Eq. 1-4 | 33 | 0.0152 | 24 | 24 | 0.0196 | 40 | 50 | 0.0192 | 39 | 54 | 0.0182 | 33 |
| ASCE 41-06 | 33 | 0.0146 | 24 | 9 | 0.0215 | 47 | 18 | 0.0217 | 68 | 33 | 0.0202 | 48 |

Table 4 shows that the most accurate results were obtained with the neural network followed by the GPR model (highlighted in orange). Both of these are black-box approaches, which makes them difficult to implement under current standards. Among the machine learning procedures that produce equations, accuracy was similar for MLR, PRM, and RLR. The RLR model was the most accurate for MP *a* but the least accurate for MP *b*. The PRM model was the most accurate for MP *b*, and second most accurate for MP *a*. All machine learning models presented an improvement over the ASCE 41-17 provisions (same as Equations 1 and 3 for MP *a*) and ASCE 41-06.

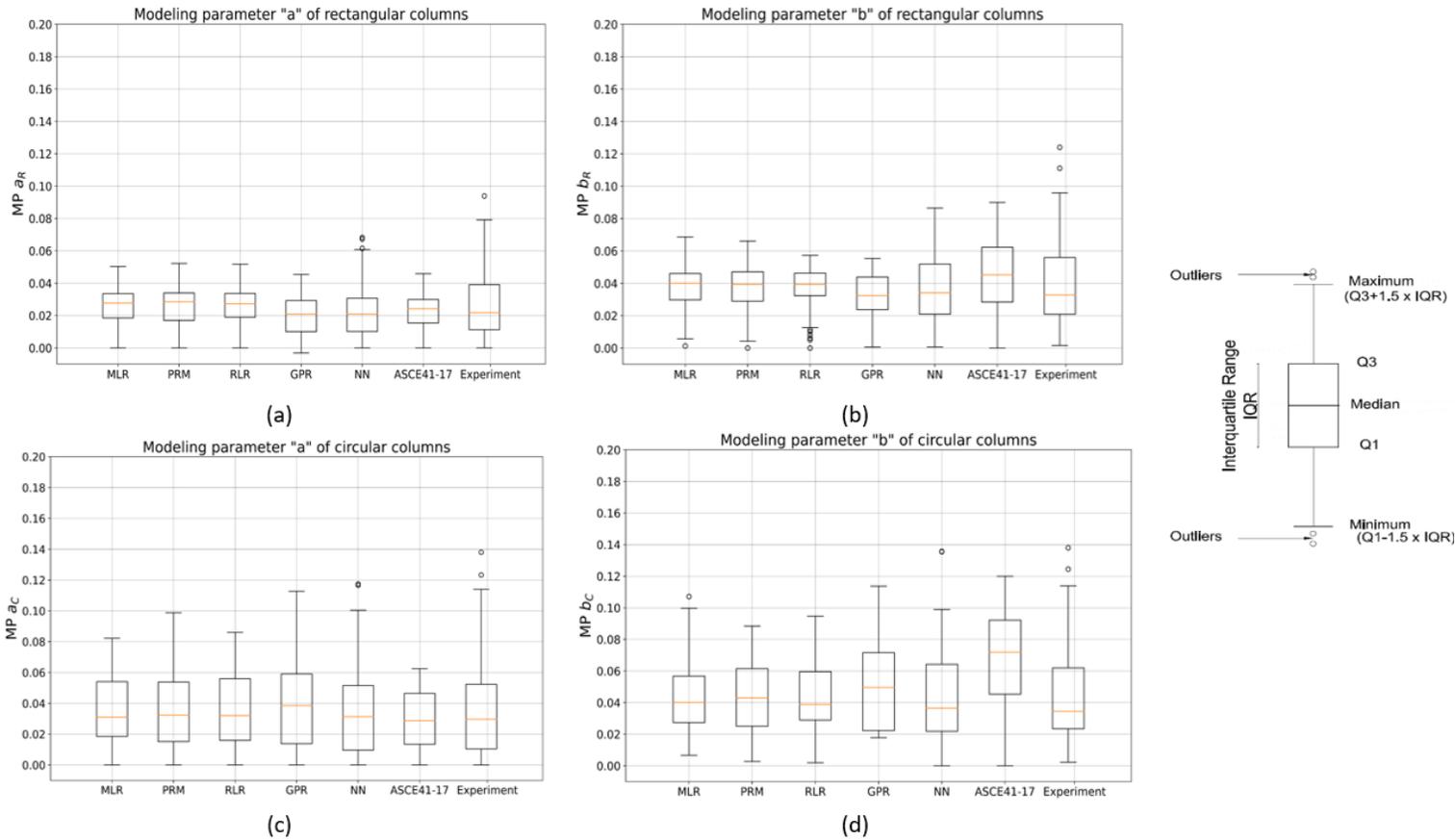

*Figure 10. Range of estimates for machine learning models evaluated.*

For all machine learning methods, the maximum defined in Figure 10 was lower than the maximum for the experimental set, for all modeling parameters (MPs). The ASCE 41-17 standard had the lowest maximum for MP *a* of rectangular and circular columns, significantly lower than the experimental set. It is important to recall that MP *a* in the ASCE 41-17 Standard is based on a linear regression model (Equations 1 and 3). The behavior is significantly different for MP *b* of rectangular and circular columns, which are based on a mechanistic model [11] had higher maxima and interquartile range upper bound than all other methods evaluated, and in most cases higher than the experimental data set. In all cases, among outliers larger than the maximum defined in Figure 10, the experimental set had the largest estimates of MPs.

For all rectangular column MPs the experimental set had a larger interquartile range than all methods evaluated. For circular column, NN and GP produced slightly larger interquartile ranges than the experimental set, and larger than all other machine learning methods. For the modeling parameter *b* in both rectangular and circular columns, the ASCE 41-17 Standard, calibrated based on the Elwood Moehle mechanistic model, had mean and interquartile ranges higher than the experimental set and all machine learning models. For median and interquartile ranges of all MPs, the NN model provided the closest match to the experimental set.

Figure 10 shows that none of the machine learning methods evaluated, including the black box approaches, produced outliers beyond measured values. For example, for rectangular columns, estimates of MP *a* obtained with the NN model were below 7 radians and 5 radians for all other machine learning models, both of which are below the maximum for the experimental set. This trend was similar across all modeling parameters and shows that the concern of black-box approaches producing unreasonably large estimates of deformation capacity did not materialize in this study.

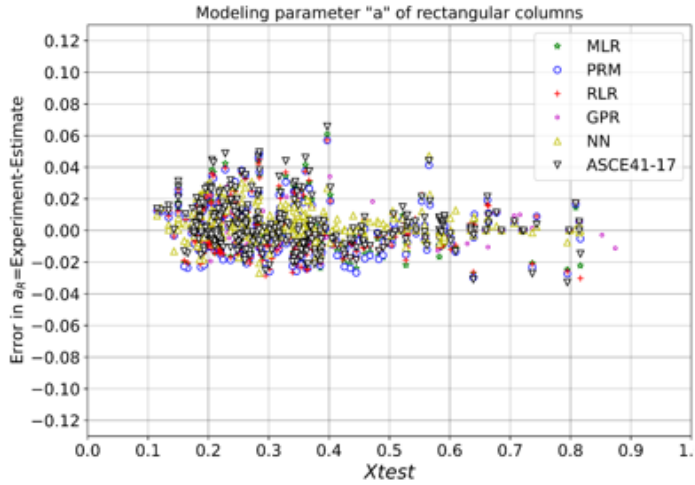
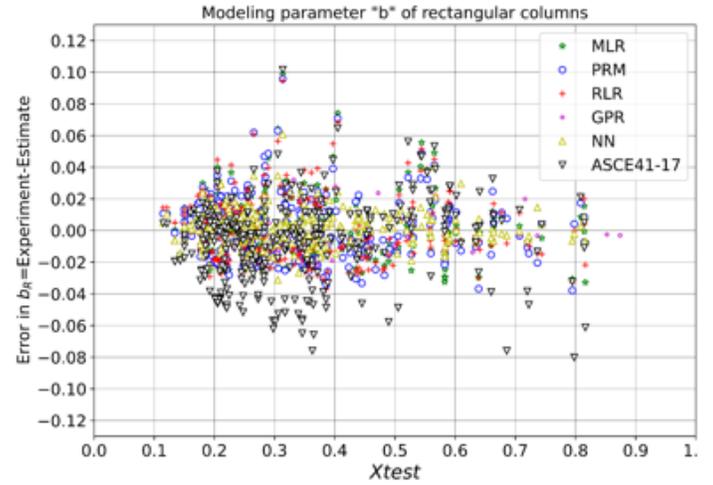

(a) (b)

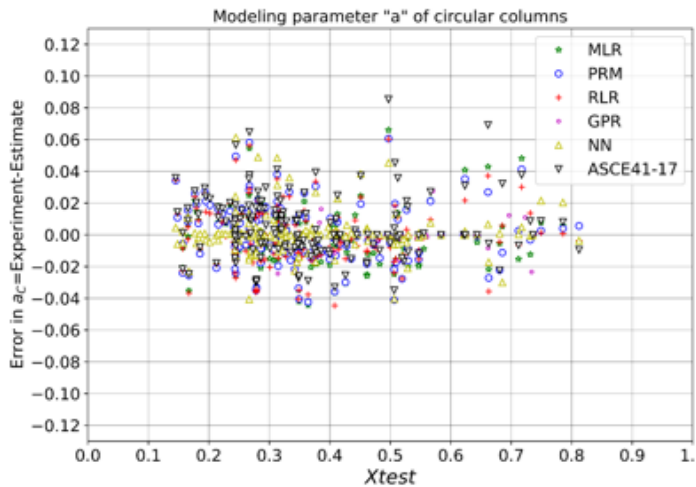
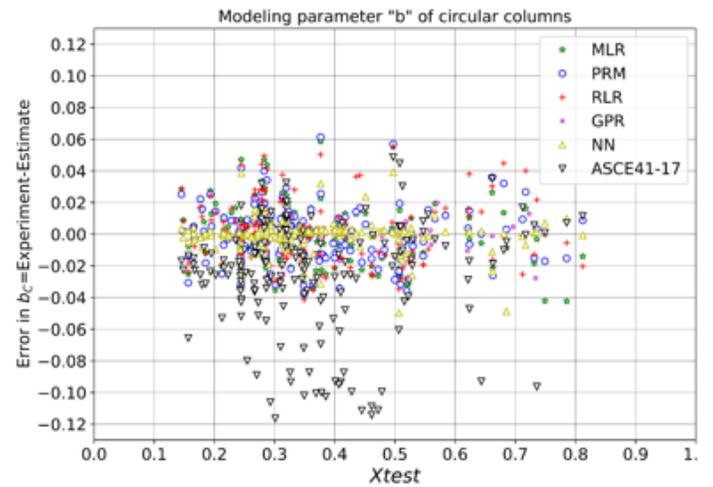

(c) (d)

*Figure 11. Error distribution as a function of separation from parameter mean.*

A significant concern with machine learning models is that the greatest confidence in the model lies on estimates near the mean of the input parameters. To gauge the accuracy of machine learning models a parameter was defined to quantify separation from the mean of the data set. The parameter was defined by taking each of the six inputs and calculating the mean and the range, defined as the difference between the maximum and minimum of each parameter, for all specimens in the set. For each specimen in the database the separation parameter *xtest* was calculated as:

$$Xtest = \frac{\sqrt{\sum d_i^2}}{0.5 \times \sqrt{6}} \tag{15}$$

Where $d_i = \frac{(Test\ input\ variable\ i - database\ mean\ for\ input\ variable\ i)}{Database\ range\ for\ input\ variable\ i}$

Figure 11 shows the error for each specimen as a function of the separation from the mean of all parameters, *xtest*. In Figure 11 negative numbers imply unconservative estimates of modeling parameters. Figure 11 shows that none of the machine learning models exhibited an increase in error as the separation from the mean increases. For rectangular columns, Figure 11 shows that error was largest for specimens close to the mean of the data set, and decreased as separation from the mean increased. For circular columns, the distribution of errors was uniform across the range.

*Failure Mode Classification (Multinomial Logistic Regression)*

The column data set used in this study includes columns with three different failure modes: Flexure Critical (FC), Flexure-Shear Critical (FSC), and Shear critical (SC). The same failure modes are identified for both circular and rectangular columns. A classification model was fitted to the input data to determine the most likely mode of failure for new observations. Because three different classes were recognized for the mode of failure, the one-vs-all approach was employed to generate 3-binary classifier models. The most likely mode of failure is identified as the class index with the maximum probability score.

The one versus all regression model was built using the three input variables considered in ASCE 41 and in the equations for *MP*s proposed by Ghannoum and Matamoros. The Gradient descent linear regression method with a learning rate of 0.5 and 5000 iterations was used for each model fitting. The learning rate, α, was chosen based on the convergence of the cost function. As shown in Figure 12, the cost function for the circular column third class had a better convergence when using a higher learning rate of 0.5 than a smaller learning rate of 0.05.

Three equations representing the probability of occurrence of each class were derived to determine the mode of failure for a new observation. The class with the highest probability corresponds to the most likely mode of failure. Equations 15 to 17 are proposed to calculate the probability of failure of each class for rectangular columns (FC, FSC, or SC).

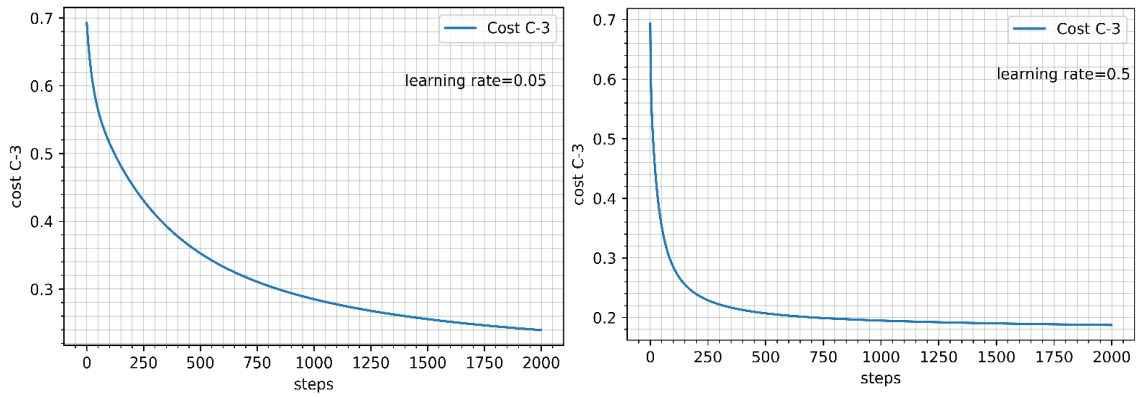

*Figure 12. The cost function for circular column third class for different learning rates.*

$$R_{FC} = 6.94 - 3.99 \frac{P}{f'_c} A_g + 0.44(\rho_t) - 9.21 \frac{V_y}{V_o} \qquad (16)$$

$$R_{FSC} = -2.19 + 0.35 \frac{P}{f'_c} A_g - 1.04(\rho_t) + 1.63 \frac{V_y}{V_o} \qquad (17)$$

$$R_{SC} = -7.7 + 4.07 \frac{P}{f'_c} A_g - 0.05(\rho_t) + 5.86 \frac{V_y}{V_o} \qquad (18)$$

The expected mode of failure corresponds to the class with the highest $R$ value among Eqs. 16 to 18. Similarly, Eqs. 19 to 21 are proposed to calculate the probability of each class for circular columns.

$$C_{FC} = 5.02 + 2.15 \left(\frac{P}{A_g f'_c}\right) - 0.2(\rho_t) - 6.35 \left(\frac{V_y}{V_o}\right) \qquad (19)$$

$$C_{FSC} = -1.52 - 3.42\left(\frac{P}{A_g f_c'}\right) + 0.02(\rho_t) + 0.8\left(\frac{V_y}{V_o}\right) \quad (20)$$

$$C_{SC} = -9.72 + 3.68\left(\frac{P}{A_g f_c'}\right) - 0.19(\rho_t) + 7.27\left(\frac{V_y}{V_o}\right) \quad (21)$$

Confusion matrices for rectangular and circular column failure modes are presented in Table 5. Cases where the forecasted mode of failure was less brittle than the observed (unconservative) are shaded in orange, and cases, where the forecasted mode of failure was more brittle than observed (conservative), are shaded in green. The accuracy of the proposed classification when only the three most significant variables were considered in the model (Eq. 15 to 19) was 79% for rectangular columns and 81% for circular columns, respectively, compared with 69% and 70% for the provisions in the ASCE 41-13 Standard. If all six input variables are considered in the classification model, accuracy increased to 87% for rectangular columns and 84% for circular columns.

The machine learning model with six variables produced failure forecasts of more ductile modes of failure (unconservative) for 10% of rectangular columns and 8% of circular columns. A forecast of less ductile modes of failure (conservative) occurred for 3% of rectangular columns and 8% of circular columns. These similarities between unconservative and conservative estimates show that the machine learning model was mostly neutral. ASCE 41-13 provisions produced failure forecasts of more ductile modes of failure (unconservative) for 8% of rectangular columns and 3% of circular columns. A forecast of less ductile modes of failure (conservative) occurred for 23% of rectangular columns and 26% of circular columns. These differences between unconservative and conservative estimates show that the ASCE 41-13 provisions were calibrated to be conservative.

Table 5 shows that the largest number of misclassifications by the machine learning model corresponded to rectangular flexure-shear critical columns misidentified as flexure-critical columns (27 columns, representing 33% of rectangular flexure-shear critical columns and 8% of the total). This was the least accurate class for both approaches evaluated. The machine learning model misclassified 40% of rectangular and 53% of circular flexure-shear critical columns, while the ASCE 41-13 provisions misclassified 43% of rectangular and 63% of circular flexure-shear critical columns. Failure modes of flexure-critical and shear-critical columns were accurately identified by both approaches. The machine learning model identified the correct mode of failure for 97% of rectangular columns and 94% of circular columns in these two classes, while the ASCE 41-13 provisions identified the correct mode of failure for 73% of rectangular columns and 80% of circular columns.

The application available on Github [33] provides the outcome of the classification model considering all the input variables.

*Table 5. Confusion matrices for rectangular and circular columns were calculated with a machine learning model with six variables and the procedure in ASCE 41-13.*

|  |  | \multicolumn{3}{c}{Rectangular Columns} | \multicolumn{3}{c}{Circular Columns} |
|---|---|---|---|---|---|---|---|
|  |  | True Class |  |  | True Class |  |  |
| Predicted Class |  | F.C. | F.S.C. | S.C. | F.C. | F.S.C. | S.C. |
|  | F.C. | 196 (144) # | 27 (13) | 1 (3) | 95 (74) | 11 (4) | 0 (0) |
|  |  | 99 (73) % | 33 (16) | 3 (8) | 95 (74) | 29 (11) | 0 (0) |
|  | F.S.C | 1 (38) | 49 (47) | 5 (9) | 4 (21) | 18 (14) | 3 (1) |
|  |  | 1 (19) | 60 (57) | 12 (22) | 4 (21) | 47 (37) | 9 (3) |
|  | S.C. | 0 (15) | 6 (22) | 34 (28) | 1 (4) | 9 (20) | 31 (33) |
|  |  | 0 (8) | 7 (27) | 85 (70) | 1 (4) | 24 (53) | 91 (97) |

The numbers and percentages in parentheses were calculated with the classification provisions in ASCE 41-13.

The main concern that arises from the use of classification algorithms is that a model may produce unconservative estimates of MPs for misclassified specimens, particularly those that are misclassified as having a more ductile mode of failure. Specific cases of specimens misclassified as having a more ductile mode of failure are shear critical specimens misclassified as flexure-shear critical, and flexure-shear critical specimens misclassified as flexure critical. Because the majority of misclassified specimens were flexure-shear critical (Table 5), a closer analysis of the error was performed for this set, and the results for the NN model and the ASCE 41 Standard are summarized in Table 6 for modeling parameter *a* of the rectangular columns.

*Table 6. Error in modeling parameter a of rectangular columns for neural network model and ASCE 41-17.*

| | | NN Error = Experiment -Estimate | | | | |
|---|---|---|---|---|---|---|
| **Observed** | **Estimated** | **Min** | **Max** | **Mean** | **Median** | **No. of specimens in the bin** |
| FSC | FSC | -0.0056 | 0.0274 | 0.0042 | 0.0034 | 49 |
| FSC | FC | -0.0078 | 0.0274 | 0.0049 | 0.0044 | 27 |
| FSC | SC | -0.0054 | 0.0059 | 0.0020 | 0.0025 | 6 |
| FC | FSC | 0.0141 | 0.0141 | 0.0141 | 0.0141 | 1 |
| SC | FSC | -0.0148 | 0.0021 | -0.0022 | 0.0011 | 5 |
| | | ASCE 41-17 Error = Experiment -Estimate | | | | |
| **Observed** | **Estimated** | **Min** | **Max** | **Mean** | **Median** | **No. of specimens in the bin** |
| FSC | FSC | -0.0127 | 0.0353 | 0.0033 | 0.0029 | 49 |
| FSC | FC | -0.0197 | 0.0311 | 0.0067 | 0.0082 | 27 |
| FSC | SC | -0.0068 | 0.0047 | -0.0018 | -0.0029 | 6 |
| FC | FSC | 0.0094 | 0.0094 | 0.0094 | 0.0094 | 1 |
| SC | FSC | -0.0217 | 0 | -0.0142 | -0.0211 | 5 |

In Table 6 unconservative misclassifications are highlighted in orange and conservative misclassifications are highlighted in green. Among misclassified flexure-critical specimens, the largest number corresponded to flexure-shear critical specimens misclassified as flexure critical, which is not surprising because it is a difficult distinction to make. Table 6 shows that for this subset of specimens the mean error was conservative and that the largest unconservative (minimum) error was less than 1%. The ASCE 41-17 Standard had both larger conservatism in terms of the mean error and the largest unconservative (minimum) error of nearly 2%.

For shear-critical specimens misclassified as flexure-shear critical, the NN model had unconservative mean and minimum errors, and both were lower than the corresponding values for flexure-shear critical specimens classified as flexure critical. In both instances, the performance of the NN model was better than the provisions in the ASCE 41-17 Standard.

The distribution of error for misclassified specimens is also illustrated in Figure 13, which shows error vs the normalized class score for flexure-shear critical specimens misclassified as flexure critical (16a) and shear critical specimens misclassified as flexure-shear critical (16b). The magnitude of the abscissa is proportional to the error in classification, and scores below one correspond to correct classification in Figure 13a and incorrect classification in Figure 13b. Figure 13a shows that most flexure-shear critical specimens misclassified as flexure critical had normalized classification scores close to one, which means that the misclassification error was small and that the range of errors was much smaller for the NN model than the provisions in the ASCE 41 Standard. Figure 13b shows that the NN model had nearly zero error for most of the misclassified specimens and that the NN performed better than the ASCE 41 Standard provisions for this set as well.

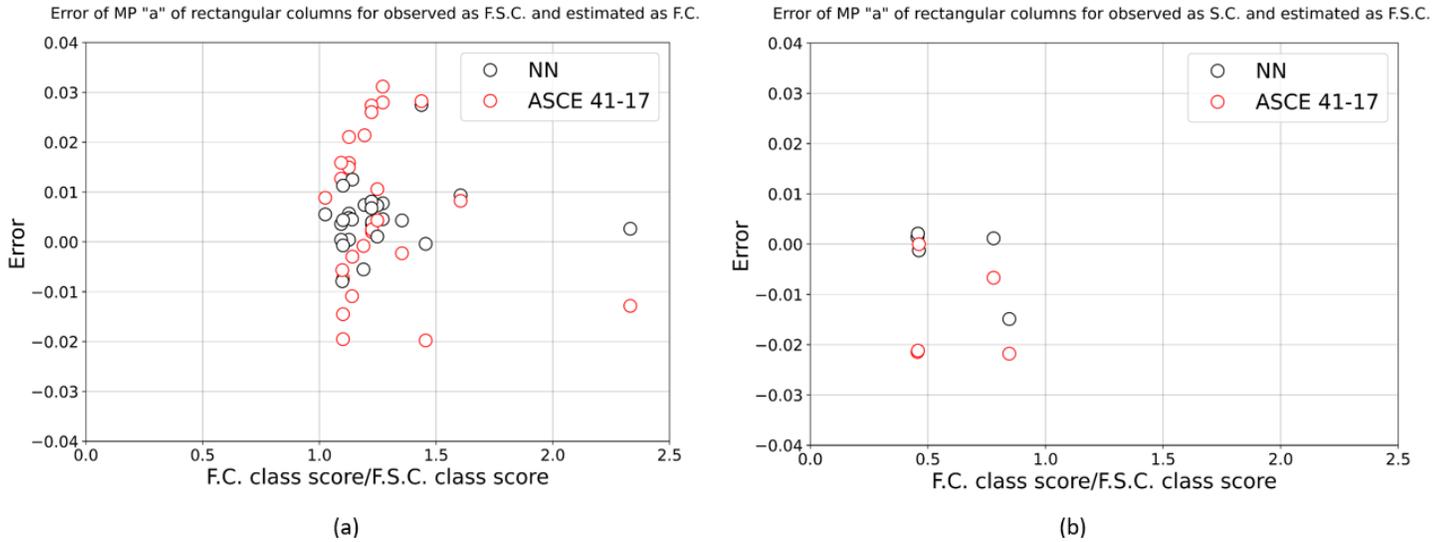

*Figure 13. Error distribution for misclassified specimens.*

## *Summary and Conclusions*

Based on the results shown in Table 3, considering only the three most significant variables, the estimation of the predicted values has improved compared to the based models, which are proposed estimates by Ghannoum and Matamoros [5] and ASCE estimates [3]. Also, the equations presented for calculating the MPs are simple enough to be used efficiently. Adding higher order of the significant variables will solve the underfitting problem of the basic model.

The regularized linear regression model and Polynomial of the three most significant variables model will offer even better estimates in terms of R-squared and Standard deviation and offer relations for predicting the modeling parameters. The NN and the Gaussian Process Regression models have the most accurate estimates, however, they do not offer simple equations for estimating the modeling parameters. After the NN and the Gaussian Process Regression models, the Polynomial of the three most significant variables will provide the most accurate and precise results.

The results of this study indicate that our NN model outperforms other methods and could be used as the baseline in future works to develop relations for the prediction of modeling parameters. Specific optimization on hyperparameters of the NN model has been performed for the circular columns models. The results of the parametric variation study are used to introduce new features to the NN model.

In addition, a classification model is presented in this study. There are three different failure modes, including Flexure Critical (FC), Flexure-Shear Critical (FSC), and Shear critical (SC) for both rectangular and circular columns. Using one versus all multinomial logistic regression, the proposed model can precisely predict the failure mode of a new observation. The Accuracy of the model is 79% for rectangular columns and 81% for circular columns. A GUI script is also designed and submitted on GitHub [33] that provides the estimates of the modeling parameters based on the NN model and the failure mode of the columns based on the Classification model. Also, a web service has been designed to make it even easier for users to have the NN and the classification model results [34].